# Automated detection of underdiagnosed medical conditions via opportunistic imaging

**Asad Aali** **Andrew Johnston** **Louis Blankemeier**
**Dave Van Veen** **Laura T Derry** **David Svec**
**Jason Hom*** **Robert D. Boutin*** **Akshay S. Chaudhari***

**Stanford University**

**Abstract**

*Abdominal computed tomography (CT) scans are frequently performed in clinical settings. Opportunistic CT involves repurposing routine CT images to extract diagnostic information and is an emerging tool for detecting underdiagnosed conditions such as sarcopenia, hepatic steatosis, and ascites. This study utilizes deep learning methods to promote accurate diagnosis and clinical documentation. We analyze 2,674 inpatient CT scans to identify discrepancies between imaging phenotypes—characteristics derived from opportunistic CT scans—and their corresponding documentation in radiology reports and ICD coding. Through our analysis, we find that only 0.5%, 3.2%, and 30.7% of scans diagnosed with sarcopenia, hepatic steatosis, and ascites (respectively) through either opportunistic imaging or radiology reports were ICD-coded. Our findings demonstrate opportunistic CT's potential to enhance diagnostic precision and accuracy of risk adjustment models, offering advancements in precision medicine.*

## Introduction

Body composition analysis is a crucial component of clinical assessments, helping identify health risks associated with obesity, malnutrition, and other conditions. Over 85 million computed tomography (CT) scans are performed each year, providing diagnostic insights into patient health [1]. The widespread use of abdominal CT scans in clinical settings indicates a vast repository of underutilized imaging data. Abdominal CT scans offer an opportunity to extract quantitative body composition metrics, serving as diagnostic and prognostic biomarkers. While manual extraction of these metrics is labor-intensive, opportunistic CT provides an automated alternative. Opportunistic CT involves leveraging existing CT scans to generate additional diagnostic insights beyond their original purpose, potentially identifying underdiagnosed conditions [2–4] including: 1) sarcopenia, which is characterized by the progressive loss of skeletal muscle mass and strength, and is associated with increased risks of falls, fractures, and mortality, 2) hepatic steatosis, which is marked by excessive fat accumulation in the liver, and is linked to metabolic disorders and cardiovascular risks, and 3) ascites, which is the pathological accumulation of fluid in the abdominal cavity, and is often indicative of a poor prognosis, particularly in patients with advanced liver disease.

Such conditions are clinically significant due to their contribution to increased mortality, particularly in aging populations [5]. Despite their clinical relevance, these conditions are often under-recognized in electronic health records (EHRs), leading to delays in diagnosis and treatment [6,7]. Deep learning offers promising advancements in body composition analysis. For example, Comp2Comp [8], a library for extracting clinical insights from CT scans has shown promise in quantifying critical body composition metrics. In parallel, vision-language foundational models like Merlin [9] have shown immense potential in accurate disease detection using CT scans. While manual assessment of muscle mass, liver morphology, and fluid accumulation can provide valuable insights, it is time-consuming and prone to interobserver variability. Deep learning algorithms can facilitate efficient and accurate detection of associations between imaging biomarkers and medical conditions, potentially at a much larger scale [10]. In this study, we evaluate 2,674 inpatient CT scans to explore discrepancies between imaging findings, radiology report findings, and International Classification of Diseases (ICD) coding for sarcopenia, hepatic steatosis, and ascites.

## Methods

**Body Composition Analysis**: To detect sarcopenia and hepatic steatosis, we used an open-source Python package designed to automate body composition analysis [8]. Comp2Comp uses convolutional neural networks to segment CT images, enabling consistent and reproducible extraction of body composition metrics for large-scale analysis. In this study, three specific pipelines were utilized from Comp2Comp: 1) 2D Spine, Muscle, and Adipose Tissue Analysis in the third lumbar vertebra (L3), 2) 2D Contrast Phase Detection, and 3) 3D Analysis of Liver and Spleen.

*co-senior author

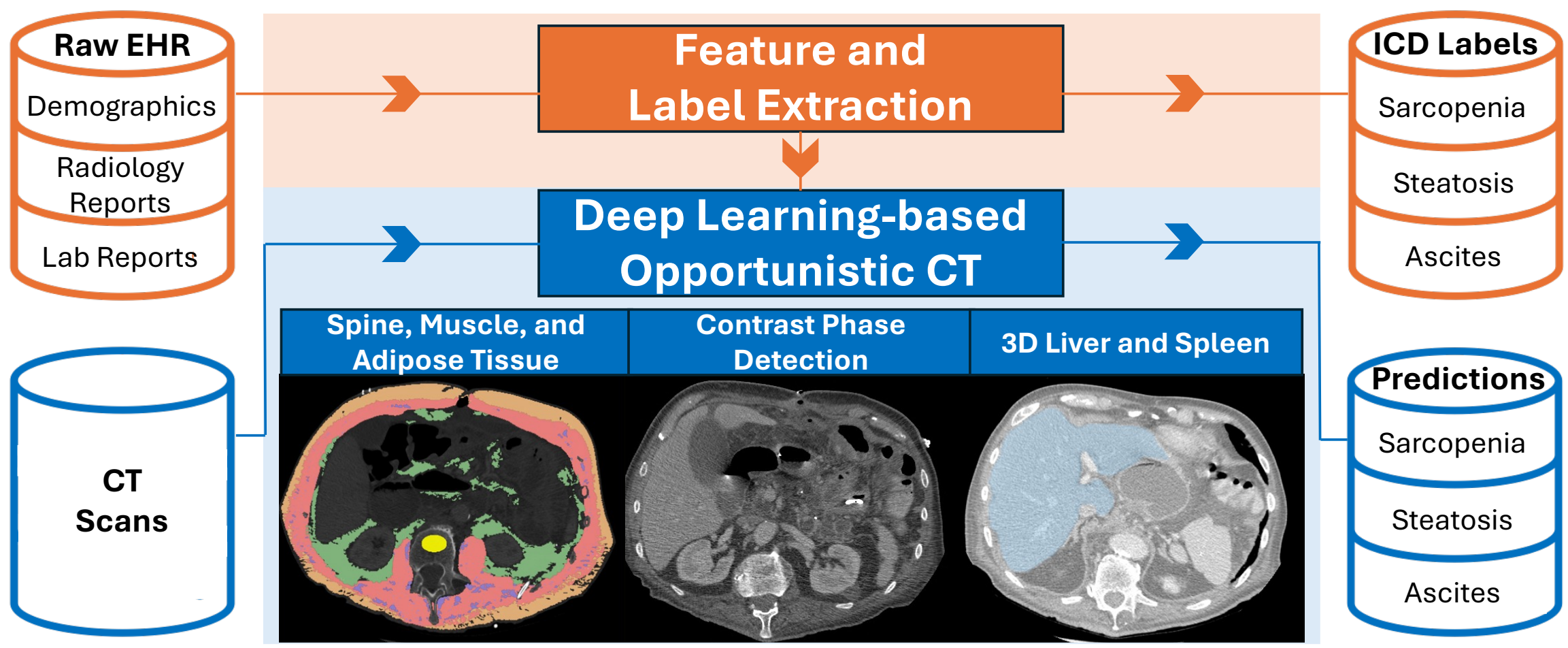

**Figure 1.** Overall study schematic: 1) We processed EHR data and extracted relevant features and ICD labels. 2) We passed EHR features and CT scans through the opportunistic CT pipeline to get disease predictions. 3) We compared predictions with radiology report diagnoses and ICD labels.

**Vision Language Foundation Model for CT**: To detect ascites, we used Merlin [9], a 3D vision-language foundation model. Originally developed to leverage structured EHR data and unstructured radiology report supervision, Merlin can process 3D volumetric data, making it suitable for abdominal CT analysis. We fine-tuned the Merlin model for ascites detection using a dataset curated by musculoskeletal radiologists, consisting of 200 labeled CT scans. We applied a train-test split of 50/50 and fine-tuned the full model weights using the Adam optimizer. The learning rate ($\lambda$) was set to $1e-5$, and we fine-tuned the model for 100 epochs.

**Data Acquisition**: This IRB-approved study included a retrospective consecutive CT dataset from a single academic medical center from 2014 to 2018, comprising 23,540 unique patients linked to 33,548 CT scans (demographics in Table 5). For each scan, the highest slice resolution series was selected. We focused exclusively on inpatient CT scans to include more severe cases. We utilized various EHR sources, including demographics, admission/discharge/transfer (ADT) records, radiology reports, and labs (Figure 1) using an internal data linkage tool. For feature and label extraction, we utilized an open-source tool called STARR-Labeler [11], resulting in a cohort containing relevant EHR features, CT scans, ICD-9 and ICD-10 labels. For each disease, we searched for relevant ICD codes within a window of 7 days before and 30 days after the scan date. After removing erroneous entries (e.g., outliers, missing values), the final cohort comprised 2,674 CT scans. For sarcopenia, we utilized the entire 2,674 inpatient cohort. For hepatic steatosis, we selected 2,275 contrast-enhanced scans, recognizing that the presence of a contrast agent can affect liver density measurements. After running Comp2Comp's contrast detection algorithm, we found that our cohort comprised 2,275 (85%) contrast-enhanced scans in the *venous* phase. While non-contrast CT scans have previously been used to measure liver density to detect fat content, they were a small portion of our cohort and were therefore excluded from the study. For ascites, we excluded scans with erroneous model outputs (null values), leaving us with 2,320 scans.

**Criteria for Clinical Diagnosis**: Sarcopenia was assessed using skeletal muscle cross-sectional area (SMA) and skeletal muscle index (SMI) (incorporating both muscle mass and radiodensity) measurements at the L3 vertebral level. We chose two independent cutoffs to validate the outcomes. The first study [12] defined a T-Score approach based on the SMI value at the L3 vertebral level (L3SMI): $\frac{L3SMI-47.5}{6.6}$ and $\frac{L3SMI-60.9}{6.6}$ for female and male patients, respectively. The second study [13] used a BMI-adjusted Z-Score (BMI-Z-Score) approach for body size adjustment based on the SMA value at L3 (L3SMA). The BMI-Z-Score is calculated as $\frac{I-\hat{I}}{SD(I)}$, where $I = \frac{L3SMA}{height}$, $\hat{I} = 50 + BMI + 13 \times sex + 0.6 \times BMI \times sex$, and $SD(I) = 8.8 + 2.6 \times sex$. In this study, scans with scores 2.5 standard deviations below the mean of the reference population were given a sarcopenia diagnosis. For hepatic steatosis, we did not find a single consensus criterion. Hence, we employed two independent Hounsfield Unit-based (HU) cutoffs: 1) liver attenuation ≤ 90 HU (achieved 75.9% and 95.7% sensitivity and specificity, respectively [14]), and 2) liver-spleen attenuation difference ≤ -19 HU (achieved 69.2% and 95.8% sensitivity and specificity, respectively [15]). We used the fine-tuned Merlin model for ascites because the volumetric assessment of intraabdominal fluid was not incorporated into our pipeline. We also analyzed radiology reports to enhance confidence in diagnosing steatosis and ascites. To detect hepatic steatosis using radiology reports, we searched for the following terms: steatosis, liver hypoattenuation, MASLD, non-alcoholic fatty liver, NAFLD, focal fatty sparing, and fatty liver. For ascites, we searched for the following terms: free fluid, free air, pelvic fluid, paracolic gutter, and loculated appearing fluid.

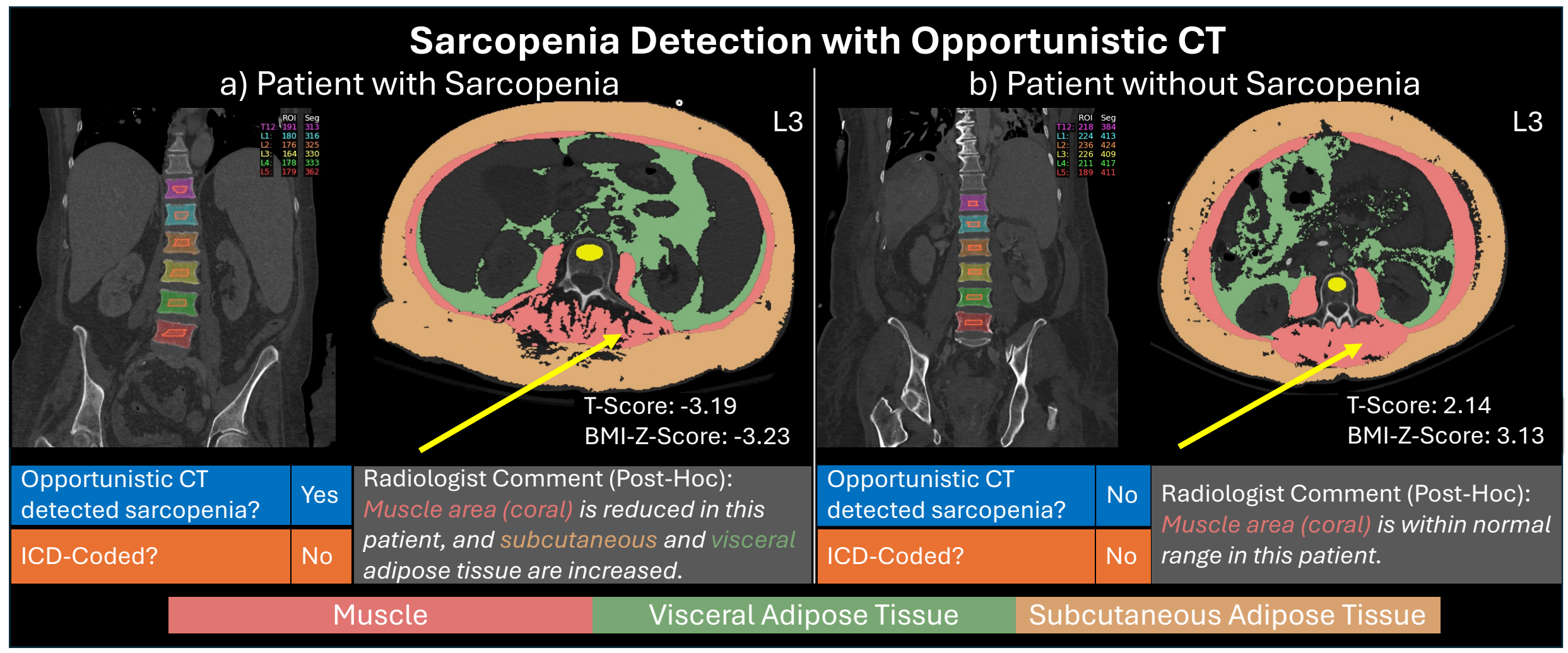

**Figure 2.** Sarcopenia - Opportunistic detection and ICD coding for: a) 66-year-old female with a BMI of 22.0, and b) 64-year-old female with a BMI of 49.8.

## Results

**Sarcopenia**: Out of the 2,674 scans, only 7 scans were coded for sarcopenia. We further identified that 388 of remaining 2,667 scans (14.6%) had T-Scores ≤ −2.5. Using BMI-Z-Scores, we identified 515 of 2,667 scans (19.3%) with scores ≤ −2.5, clearly representing the scale of sarcopenia under-coding. Of the 7 scans that were coded for sarcopenia, only 1 (14.3%) and 3 (42.9%) scans had T-Scores and BMI-Z-Scores ≤ −2.5 (possible over-coding). Even in cases where both T-Score and BMI-Z-Score approaches agreed (302 scans) and had scores ≤ −2.5, only 1 (0.33%) case was ICD-coded. The overlap between T-Score and BMI-Z-Score approaches and ICD coding is shown in Table 1. In Figure 2, we show two examples (not sarcopenia coded) with qualitative comments from a radiologist in our team, where cases 1 and 2 show patients with/without sarcopenia. Figure 5 shows the distribution of T-Scores and BMI-Z-Scores, categorized by ICD diagnosis.

**Table 1**: Overlap in sarcopenia detection using: a) T-Scores, b) BMI-Z-Scores, c) ICD coding.

| T-Score ≤ -2.5 | BMI-Z-Score ≤ -2.5 | ICD-Coding | Count | % |
|---|---|---|---|---|
| Yes | Yes | Yes<br>No | 1<br>301 | 0.0%<br>11.3% |
| Yes | No | Yes<br>No | 0<br>87 | 0.0%<br>3.3% |
| No | Yes | Yes<br>No | 2<br>214 | 0.1%<br>8.0% |
| No | No | Yes<br>No | 4<br>2,065 | 0.1%<br>77.3% |
| **Total** | | | **2,674** | **100.0%** |

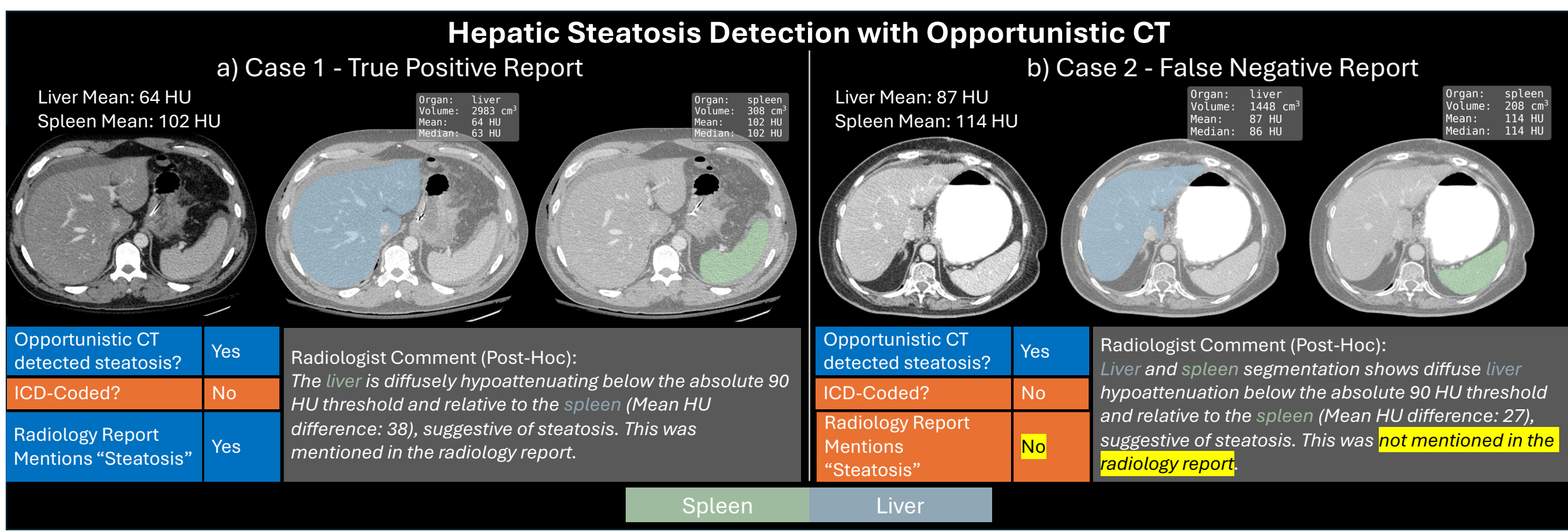

**Figure 3.** Hepatic steatosis - Opportunistic detection, radiology report diagnosis, and ICD coding for: a) 46-year-old male with a BMI of 31.25, and b) 68-year-old female with a BMI of 17.37.

**Hepatic Steatosis**: Out of the 2,275 contrast-enhanced scans, only 28 scans were coded for hepatic steatosis. From the remaining 2,247 scans, we identified that 227 (10.1%) scans had a liver attenuation of ≤ 90 HU. Similarly, 370 of the 2,247 scans (16.5%) had a liver-spleen attenuation difference of ≤ −19 HU, representing the scale of under-coding. Of the 28 scans that were coded for steatosis, only 7 (25%) and 14 (50%) scans got an opportunistic diagnosis based on the liver and liver-spleen approaches, respectively (possible over-coding). Even in cases where both cutoff methods agreed on steatosis diagnosis (132 scans), only 6 (4.5%) cases were ICD-coded. Furthermore, our analysis showed 63 scans where both quantitative approaches and radiology reports agreed on a steatosis diagnosis. From these 63 scans, only 5 (7.9%) were ICD-coded. The overlap between the two independent quantitative approaches, radiology report diagnoses, and ICD coding is detailed in Table 2. In Figure 3, we show two cases from our cohort with observations from our team radiologist. Case 1 shows a hepatic steatosis example that was correctly diagnosed through opportunistic imaging and in the radiology report but was not coded. While case 2 was diagnosed through opportunistic imaging, it was neither coded, nor mentioned in the radiology report, demonstrating the benefit of opportunistic CT beyond coding and radiology reports. Figure 6 shows the distribution of mean liver HU and liver-spleen difference HU, categorized by ICD diagnosis.

**Table 2**: Overlap in steatosis detection using: a) Liver HU, b) Liver-Spleen HU, c) Radiology Reports, d) ICD coding.

| Liver ≤ 90 HU | Liver-Spleen ≤ -19 HU | Radiology Report | ICD-Coding | Count | % |
|---|---|---|---|---|---|
| Yes | Yes | Yes | Yes | 5 | 0.2% |
| | | | No | 58 | 2.5% |
| | | No | Yes | 1 | 0.0% |
| | | | No | 68 | 3.0% |
| Yes | No | Yes | Yes | 1 | 0.0% |
| | | | No | 16 | 0.7% |
| | | No | Yes | 0 | 0.0% |
| | | | No | 85 | 3.7% |
| No | Yes | Yes | Yes | 2 | 0.1% |
| | | | No | 33 | 1.5% |
| | | No | Yes | 6 | 0.3% |
| | | | No | 211 | 9.3% |
| No | No | Yes | Yes | 2 | 0.1% |
| | | | No | 52 | 2.3% |
| | | No | Yes | 11 | 0.5% |
| | | | No | 1,724 | 75.8% |
| **Total** | | | | **2,275** | **100.0%** |

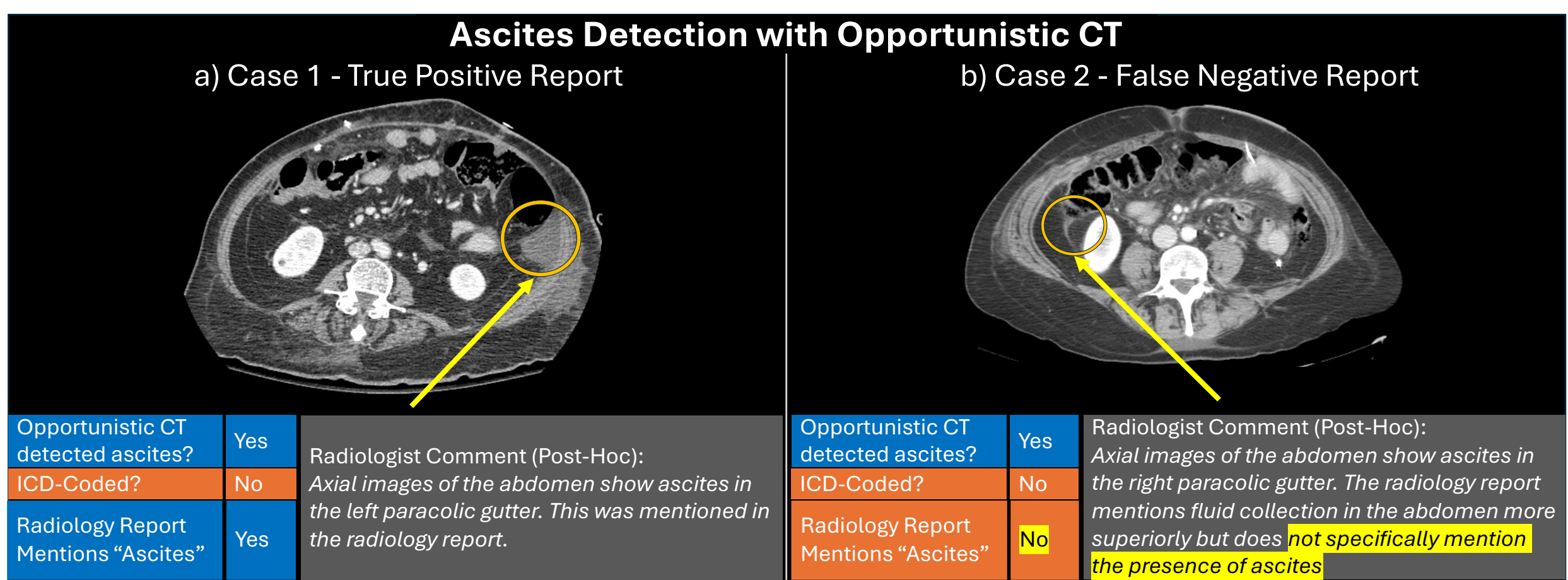

**Figure 4.** Ascites - Opportunistic detection, radiology report diagnosis, and ICD coding for: a) 68-year-old female with a BMI of 17.37, and b) 68-year-old female with a BMI of 22.58.

**Ascites**: Ascites detection was performed using sigmoid outputs from the fine-tuned Merlin model. For predicting ascites, we optimized the classification threshold given the test results in Table 4. We chose a threshold of 0.6, prioritizing higher *precision* and *specificity* while maintaining a minimum F1-Score of 70%. Our evaluation of the Merlin model relied on the fine-tuned model's validated performance, with no further manual verification. From the 2,320 scans, 644 scans were coded for ascites. Of the remaining 1,676 scans not coded, the fine-tuned Merlin model diagnosed 1,101 (65.7%) scans with ascites. Furthermore, from the 1,676 scans not coded for ascites, we found 696 (41.5%) scans diagnosed through opportunistic imaging and the radiology reports, representing the under-coding scale. Furthermore, we observed that from the 644 scans that were coded for ascites, only 372 (57.8%) scans got a diagnosis through opportunistic imaging and radiology reports, possibly suggesting over-coding. We also report that out of all cases where Merlin predicted ascites, the radiology report diagnosis agreed 66.3% of times. However, the radiology report diagnoses proportion dropped to 33.6% in cases where the model did not predict ascites, representing a positive correlation between the two approaches. The overlap between the Merlin model's predictions, radiology report diagnoses, and ICD coding is further detailed in Table 3. In Figure 4, we show two cases from our cohort (not ascites coded) with comments from our team radiologist. Case 1 shows an example of ascites that was correctly diagnosed through opportunistic imaging and radiology reports but was not coded. While case 2 was diagnosed through opportunistic imaging, it was neither coded nor mentioned in the radiology report.

**Table 3**: Overlap in ascites detection using: a) Merlin Predictions, b) Radiology Reports, c) ICD coding.

| Model Prediction | Radiology Report | ICD-Coding | Count | % |
|---|---|---|---|---|
| Yes | Yes | Yes | 372 | 16.0% |
| | | No | 696 | 30.0% |
| | No | Yes | 139 | 6.0% |
| | | No | 405 | 17.5% |
| No | Yes | Yes | 56 | 2.4% |
| | | No | 182 | 7.8% |
| | No | Yes | 77 | 3.3% |
| | | No | 393 | 17.0% |
| **Total** | | | **2,320** | **100.0%** |

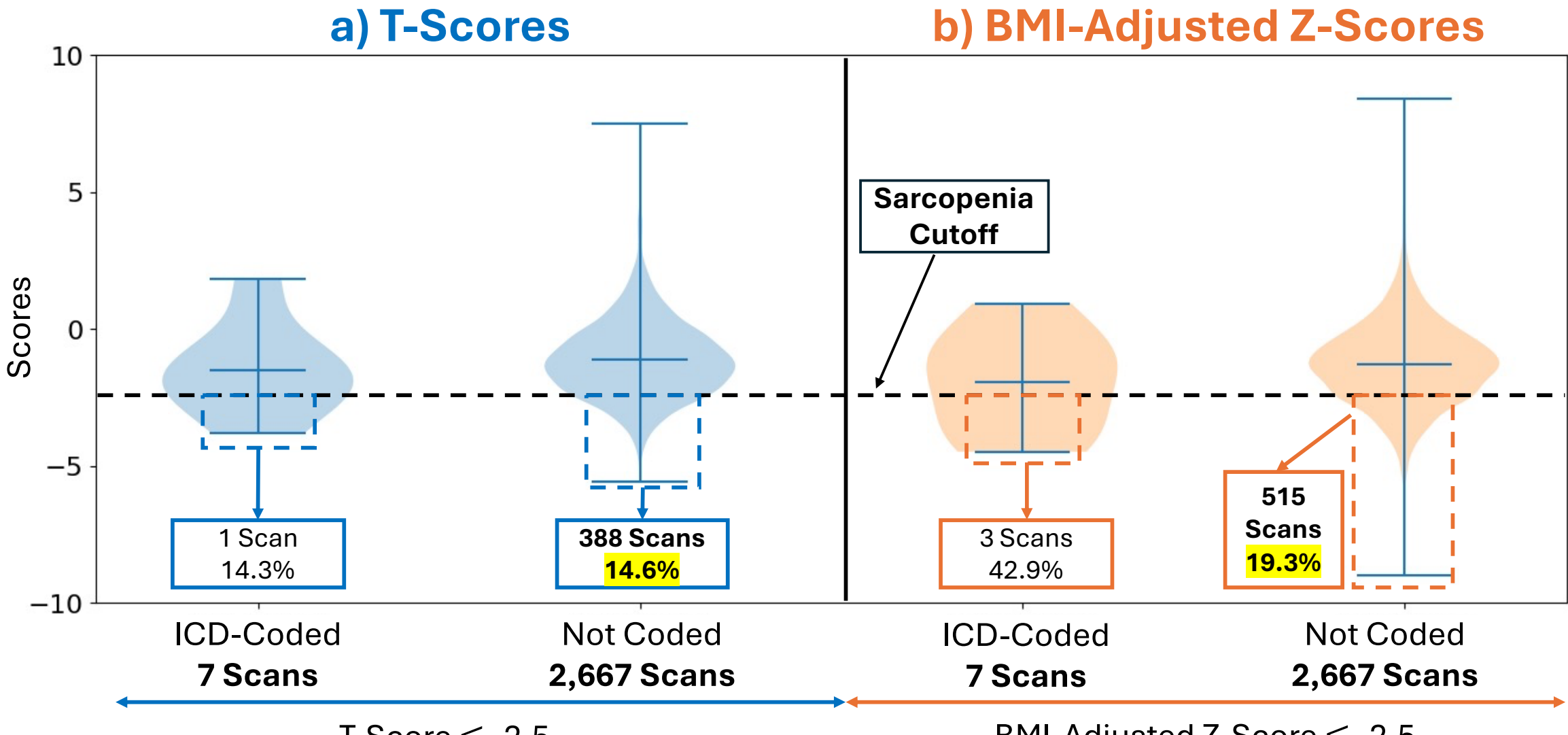

**Figure 5.** Sarcopenia detection - Distribution of a) T-Scores and b) BMI-Z-Scores.

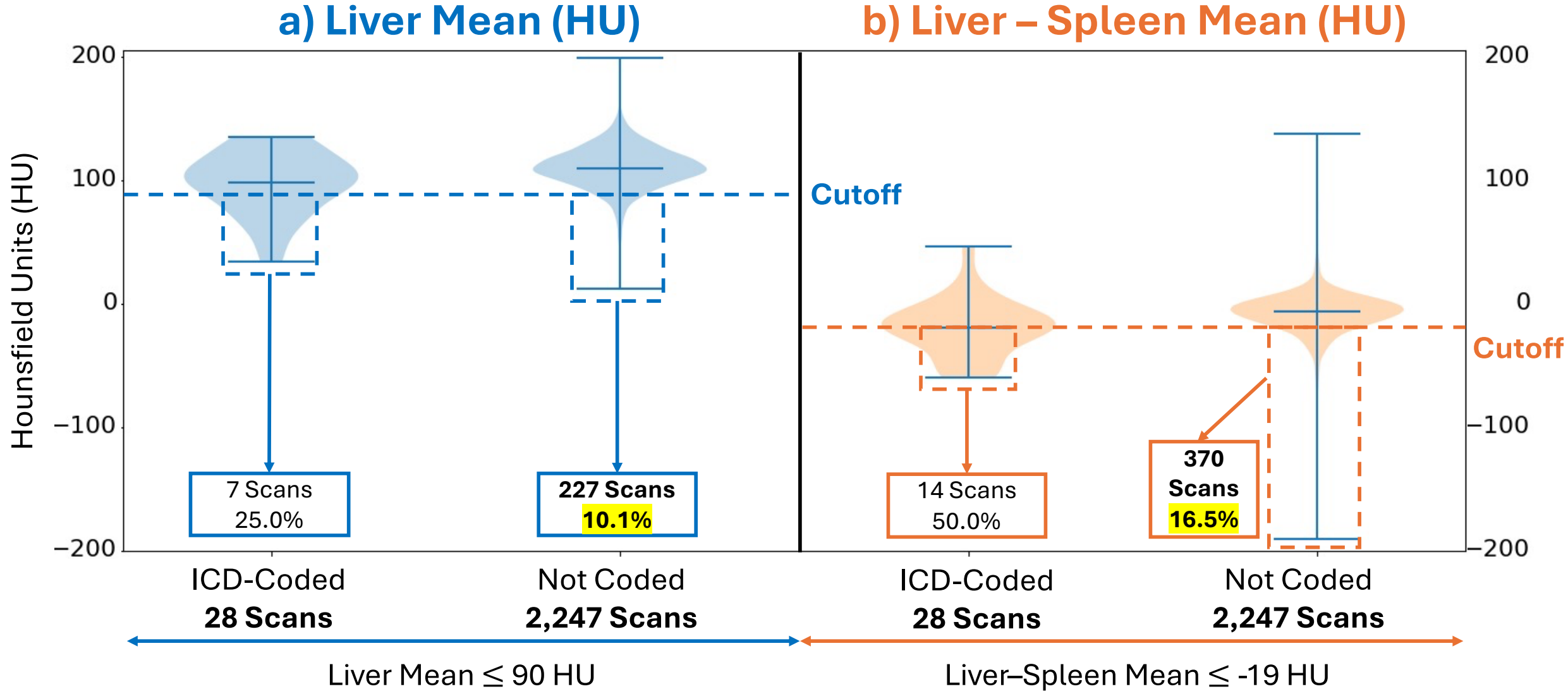

**Figure 6.** Hepatic steatosis detection - Distribution of a) liver HU and b) liver-spleen HU.

**Table 4.** Ascites detection - Test performance metrics for the fine-tuned Merlin model (N=100).

| Prediction Threshold | Precision | Sensitivity | Specificity | F1-Score | Accuracy | AUROC |
|---|---|---|---|---|---|---|
| 0.4 | 77.4% | **96.0%** | 71.4% | **85.7%** | **83.8%** | **90.3%** |
| 0.5 | 84.1% | 74.0% | 85.7% | 78.7% | 79.8% | |
| 0.6 | 87.9% | 58.0% | 91.8% | 69.9% | 74.7% | |
| 0.7 | **95.0%** | 38.0% | **98.0%** | 54.3% | 67.7% | |

**Table 5**: Demographic, sex, ethnicity, and race distribution of the extracted inpatient cohort from EHR data (N = 2,674).

| Age | Height (cm) | BMI | Male / Female (%) | Non-Hispanic / Hispanic/Latino / Other (%) |
|---|---|---|---|---|
| 57.3 ± 15.9 | 167.9 ± 10.9 | 26.2 ± 6.6 | 53%/47% | 77%/22%/1% |

| White | Other | Asian | Black | Unknown | Pacific Islander | Native American |
|---|---|---|---|---|---|---|
| 1, 453 (54%) | 651 (24.3%) | 361 (13.5%) | 116 (4.3%) | 47 (1.8%) | 24 (0.9%) | 22 (0.8%) |

## Discussion

This study utilized deep learning-driven models—Comp2Comp for detecting sarcopenia and hepatic steatosis, and the Merlin model for ascites prediction—to assess the prevalence of these conditions in 2,674 inpatient CT scans. Discrepancies were identified between sarcopenia diagnoses and ICD coding, with only 0.5% scans being ICD-coded out of all scans diagnosed with sarcopenia via either the T-Score or BMI-Z-Score criteria. Similarly, of the scans diagnosed with hepatic steatosis via either the mean liver HU, liver-spleen HU criteria, or radiology reports, only 3.2% scans were ICD-coded. Finally, of the scans diagnosed with ascites through radiology reports or opportunistic imaging, only 30.7% scans were ICD-coded. While our study focused on analyzing potential under-coding, over-coding discrepancies were also found across the three medical conditions. For example, 16.4% of cases that showed neither a mention of ascites in radiology reports nor were classified positive by the Merlin model, were ICD-coded for ascites. These findings indicate the need for further investigation into such cases to refine both deep learning models and clinical coding practices. The threshold of the outputs of deep learning models like Merlin can also be optimized for diverse clinical applications.

The ability to accurately diagnose sarcopenia, hepatic steatosis, and ascites through opportunistic CT has significant clinical implications. Early identification of these conditions enables timely interventions that can improve patient outcomes. For example, diagnosing sarcopenia early allows for dietary interventions and lifestyle changes that can mitigate muscle loss and associated risks such as falls and fractures. Similarly, detecting hepatic steatosis earlier can facilitate weight management strategies, including emerging treatments like Ozempic, potentially preventing progression to more severe liver diseases. In the case of ascites, identifying the underlying cause—often cirrhosis—can enable targeted investigations and treatments, reducing complications and improving patient quality of life. Furthermore, employing more reliable large-scale studies through deep learning-driven models can improve diagnostic accuracy, and ultimately streamline clinical workflows.

However, this study has limitations, including its retrospective nature and reliance on a single institution inpatient dataset, which can affect generalizability. Additionally, integrating imaging and EMR data across multiple sites presents significant challenges due to variations in data standards, infrastructure, and coding practices, which may limit the broader applicability of these findings. The use of ICD codes as a reference standard could also introduce bias due to inconsistencies in documentation and coding practices, particularly in the inpatient setting where coders cannot directly code from radiology reports, unlike in the outpatient setting [16]. Future studies should include prospective evaluations, validation across multiple institutions, and application to additional diseases with long-term health impacts.

## Conclusion

This study demonstrates the potential of deep learning-based opportunistic CT in improving the detection and coding of medical conditions. Accurate capture of diagnoses is crucial, as incomplete data can result in suboptimal patient management, inaccurate billing, and skewed research. Integrating opportunistic imaging can enhance diagnostic accuracy while minimizing under-diagnosis and over-diagnosis. Through our analysis, we demonstrate that deep learning-based opportunistic imaging tools have the potential to strengthen both clinical practice and precision medicine research.

## Acknowledgements

We would like to thank Michelle McCormack (Director of Clinical Documentation Integrity), and Marisa Samp (Director of Hospital and Professional Coding), for their valuable support and insights regarding the project. Furthermore, we would like to acknowledge research support from NIH grants: R01 HL167974, R01 HL169345, and R01 EB002524.